\documentclass[letterpaper, 10 pt, conference]{ieeeconf}  % Comment this line out if you need a4paper

\IEEEoverridecommandlockouts                              % This command is only needed if 
\usepackage{multicol}
\usepackage[colorlinks=true, linkcolor=blue, bookmarks=true, urlcolor=blue, anchorcolor=blue]{hyperref}
\usepackage{graphicx}
\usepackage{amsmath}
\usepackage{amssymb}
\usepackage{booktabs}
\usepackage{caption}
\usepackage{multirow}
\usepackage{color}
\usepackage{ulem}
\usepackage{float}
\usepackage{bbding}
\usepackage{multirow}
\usepackage{colortbl}
\usepackage{tikz}
\usepackage{chapterbib}    
\usepackage[noadjust]{cite}
\usepackage{kantlipsum}
\usepackage{cuted}
\usepackage{graphicx}
\usepackage{amsmath}
\usepackage{amssymb}
\usepackage{booktabs}
\usepackage[figuresright]{rotating}
\usepackage{subcaption}
  \usepackage{algpseudocode}

\usepackage{amsmath,amsfonts,bm}
\usepackage{amsfonts}
\usepackage{float}
\usepackage{enumerate}
\let\labelindent\relax
\usepackage{enumitem}

\usepackage{algcompatible}

\usepackage{algorithm}
\usepackage{algorithmicx}
\usepackage{algpseudocode}

\usepackage{graphicx}
\usepackage{svg}
\usepackage{wrapfig}
\usepackage{multirow}
\usepackage{booktabs}
\usepackage[export]{adjustbox}
\usepackage{wrapfig,lipsum,booktabs}
\usepackage{xspace}
\usepackage{enumerate}
\usepackage{enumitem}
\usepackage{caption}
\usepackage[export]{adjustbox}
\usepackage{algpseudocode}

\def\eqref#1{equation~\ref{#1}}
\def\1{\bm{1}}

\def\vtheta{{\bm{\theta}}}

\DeclareMathAlphabet{\mathsfit}{\encodingdefault}{\sfdefault}{m}{sl}
\SetMathAlphabet{\mathsfit}{bold}{\encodingdefault}{\sfdefault}{bx}{n}

\def\sC{{\mathbb{C}}}
\def\sD{{\mathbb{D}}}
\def\sN{{\mathbb{N}}}

\def\sR{{\mathbb{R}}}
\def\sS{{\mathbb{S}}}

\def\sX{{\mathbb{X}}}

\title{\LARGE \bf
CEFHRI: A \uline{C}ommunication \uline{E}fficient \uline{F}ederated Learning Framework for Recognizing Industrial \uline{H}uman-\uline{R}obot \uline{I}nteraction
}

\author{Umar Khalid$^{1}$, Hasan Iqbal$^{2}$, Saeed Vahidian$^{3}$, Jing Hua$^{2}$, Chen Chen$^{1}$%%, <-this % stops a space
\thanks{*This work is supported by the
NSF/Intel Partnership on MLWiNS under Grant No.
2003198.}% <-this % stops a space
\thanks{$^{1}$Center For Research in Computer Vision, University of Central Florida, Orlando, FL, USA
        {\tt\small umarkhalid@knights.ucf.edu and chen.chen@crcv.ucf.edu}}%
\thanks{$^{2}$Dept. of Computer Science, Wayne State University, Detroit, MI, USA
        {\tt\small hasan.iqbal.cs@wayne.edu and jinghua@wayne.edu}}%
\thanks{$^{3}$Dept. of Electrical and Computer Engineering, Duke University, Durham, NC, USA 
        {\tt\small saeed.vahidian@duke.edu}.}%
}

\begin{document}
\maketitle

% \maketitle
\thispagestyle{empty}
\pagestyle{empty}

%%%%%%%%%%%%%%%%%%%%%%%%%%%%%%%%%%%%%%%%%%%%%%%%%%%%%%%%%%%%%%%%%%%%%%%%%%%%%%%%
\begin{abstract}
 Human-robot interaction (HRI) is a rapidly growing field that encompasses social and industrial applications. Machine learning plays a vital role in industrial HRI by enhancing the adaptability and autonomy of robots in complex environments. However, data privacy is a crucial concern in the interaction between humans and robots, as companies need to protect sensitive data while machine learning algorithms require access to large datasets. Federated Learning (FL) offers a solution by enabling the distributed training of models without sharing raw data. Despite extensive research on Federated learning (FL) for tasks such as natural language processing (NLP) and image classification, the question of how to use FL for HRI remains an open research problem. The traditional FL approach involves transmitting large neural network parameter matrices between the server and clients, which can lead to high communication costs and often becomes a bottleneck in FL. This paper proposes a communication-efficient FL framework for human-robot interaction (CEFHRI) to address the challenges of data heterogeneity and communication costs. The framework leverages pre-trained models and introduces a trainable spatiotemporal adapter for video understanding tasks in HRI. Experimental results on three human-robot interaction benchmark datasets: HRI30, InHARD, and COIN demonstrate the superiority of CEFHRI over full fine-tuning in terms of communication costs. The proposed methodology provides a secure and efficient approach to HRI federated learning, particularly in industrial environments with data privacy concerns and limited communication bandwidth. Our code is available at \url{https://github.com/umarkhalidAI/CEFHRI-Efficient-Federated-Learning}.

% Experiments are conducted on three human-robot interaction benchmark datasets: HRI30, InHARD, and COIN under various FL settings to validate the effectiveness of the proposed parameter-efficient FL framework.

% Recent studies have shown that pre-trained models can enhance FL performance by stabilizing global aggregation in the presence of statistical data heterogeneity

\end{abstract}
%%%%%%%%%%%%%%%%%%%%%%%%%%%%%%%%%%%%%%%%%%%%%%%%%%%%%%%%%%%%%%%%%%%%%%%%%%%%%%%%

\section{Introduction}
\begin{figure*}
    \includegraphics[trim={0 8.5cm 0 0},clip,scale = 0.53]{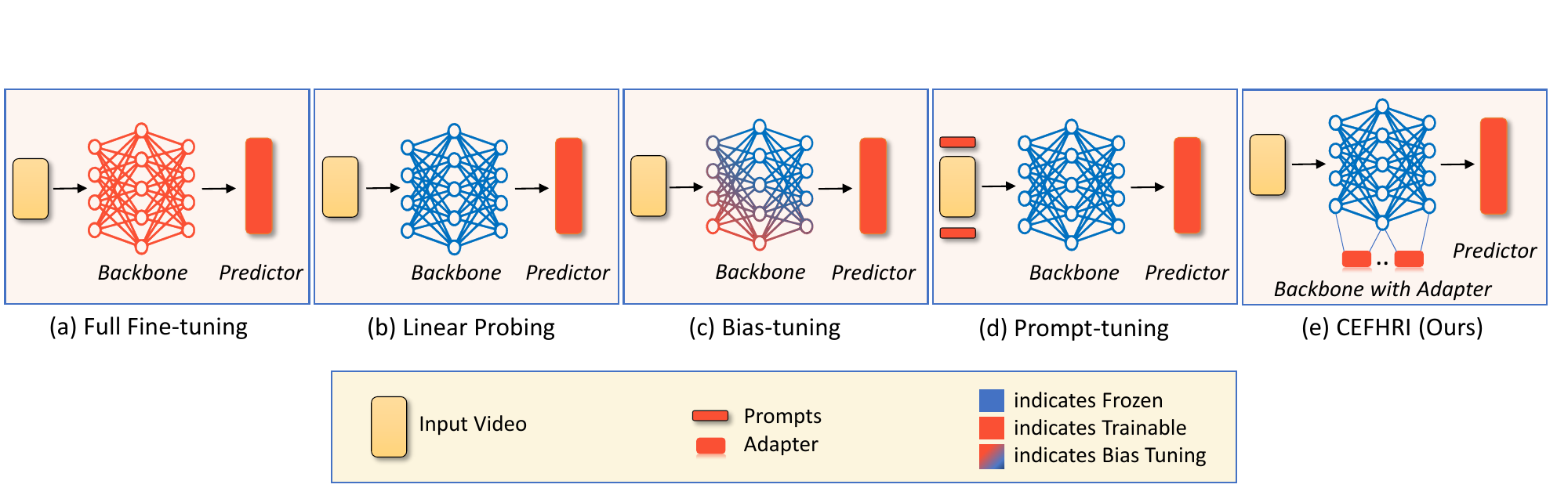}
\centering
\caption{\footnotesize Methods to fine-tune the pre-trained model (a) Full fine-tuning trains all the parameters of the pre-trained model. (b) Linear probing only fine-tunes the penultimate linear layer  (c) Fine-tunes the bias term along with the penultimate layer. (d) Prompt tokens are concatenated with the input tokens, where prompt tokens are fine-tuned with the linear classifier layer. (e) Adapters are added to each layer of the model, where only the adapters and penultimate linear layers are fine-tuned.}
\label{fig:method}
\end{figure*}
Human-robot interaction (HRI) is a rapidly growing field encompassing social~\cite{breazeal2004designing} and industrial applications~\cite{he2019industrial,9981793}. In social settings, HRI involves interactions for entertainment, education, therapy, and personal assistance~\cite{mahdi2022survey}. Conversely, Industrial HRI focuses on collaboration between humans and robots in industrial environments.
Comprehending Industrial HRI is vital for the design and utilization of secure and efficient robotic systems in industrial environments, ultimately leading to increased productivity and optimized interaction between humans and robots. In the context of Industrial HRI, machine learning plays a crucial role in facilitating the adaptability of robots to ever-changing and intricate industrial environments~\cite{choudhury2019utility}. This enhances their capability to execute tasks precisely and autonomously, while also reducing the potential for accidents and harm to human workers. 

Regarding the interaction between humans and robots in industrial settings, data privacy is an essential concern as companies need to protect sensitive data while machine learning algorithms often require access to large datasets to make accurate predictions~\cite{9857233}. Federated Learning (FL)~\cite{mendieta_local_2022} offers a solution by enabling distributed training without sharing raw data. FL ensures privacy while improving human-robot interaction and productivity. Despite its potential, FL faces challenges in remote locations with limited bandwidth, hindering communication and data transfer, impacting its widespread deployment and effective utilization \cite{Vahidian-federated-2021}. The substantial communication costs in FL due to parameter/data transmission between clients and servers also present a bottleneck.

\textbf{Our Approach.} 
To address the FL communication cost challenge, we propose a communication-efficient Federated Learning (FL) framework for Human-Robot Interaction (HRI) action recognition, named CEFHRI. We leverage pre-trained video models and fine-tune a carefully designed adapter specifically for video understanding on HRI datasets. The CEFHRI framework addresses challenges related to data heterogeneity~\cite{li2022} and large communication costs in HRI. The study is the first to investigate communication efficiency in FL for video understanding in the context of HRI.
% To address the FL communication cost challenge in the domain of HRI action recognition, we aim to introduce a \uline{c}ommunication \uline{e}fficient \uline{F}L framework for \uline{h}uman-\uline{r}obot \uline{i}nteraction (CEFHRI) that aims to limit communication costs by leveraging the pre-trained models.  
% Recent research has underscored the advantages of commencing FL training with pre-trained video models to boost performance in scenarios characterized by statistical data heterogeneity~\cite{li2022}. 
% Consequently, in our study, we opt to use the pre-trained model and then fine-tune the carefully designed adapter specifically adapted for video understanding on HRI datasets. The application of the proposed adapter has been compared with other fine-tuning baselines in Fig.~\ref{fig:method}. The proposed framework incorporating the HRI adapter, named CEFHRI, addresses the challenges posed by both data heterogeneity and large communication costs. To the best of our knowledge, this is the first work to investigate the communication efficiency of the FL in the realm of HRI as a video understanding task. 
 The key contributions of this study can be summarized as follows:
\begin{itemize}
\item This paper offers a pioneering study that systematically explores the effects of pre-training in the context of human-robot interaction (HRI) through FL.
\item The study introduces a parameter-efficient fine-tuning framework, CEFHRI, within the Vision Transformer architecture. CEFHRI addresses the challenges of data heterogeneity and large communication costs by using a lightweight trainable spatiotemporal adapter.
\item Furthermore, the proposed CEFHRI framework is evaluated for preserving model privacy on the server while achieving efficient transfer learning. 
% \item The results of comprehensive experiments conducted on three standard datasets for human-robot interaction, namely HRI30, InHARD, and COIN, demonstrate the superiority of CEFHRI over full fine-tuning with respect to the communication cost.
\item The proposed methodology is suitable for industrial environments prioritizing data privacy and facing communication and bandwidth constraints, providing a communication-efficient alternative for HRI federated learning. It can also be extended to other FL scenarios involving video understanding tasks.
\end{itemize}

\vspace{-2mm}
\section{Related Work}
\label{sec:related_work}
\paragraph{Industrial Human-Robot Interaction}
 In recent years, industrial human-robot interaction (HRI) has emerged as a prominent research field, attracting considerable interest~\cite{roitberg2014human,hentout2019human}. However, there are a limited number of studies that have explored the computer vision domain specifically action recognition in industrial HRI settings~\cite{iodice2022hri30}. 
 One study by Huang et al. \cite{huang2021convolutional} proposes a model for gesture recognition that uses Convolutional Neural Networks (CNNs) to recognize hand gestures performed by human workers in industrial environments. Rizzi et al. \cite{rizzi2019learning} explores the use of deep learning to improve the performance of robotic grasping in industrial environments, by training a model to predict the optimal grasp point for objects based on visual cues. \cite{spina2020deep} applied deep learning using a combination of convolutional and recurrent neural networks to improve the perception of robots in an industrial setting, enabling them to recognize and locate objects on a conveyor belt.
\paragraph{Efficient Fine-Tuning}
\cite{liu2018structured} proposes a novel structured pruning method for parameter-efficient fine-tuning that preserves important network structures while discarding unimportant connections.~\cite{Transfer-Learning-for-NLP-2019} realizes transfer learning in NLP tasks with adapter modules, which adds a few trainable parameters per task while keeping the backbone frozen. Some recent works ~\cite{yang2023aim,pan_parameter-efficient_2022, chen2022adaptformer} extend the adapter design to use the image foundation models for video understanding. \cite{prompt2021, zhang2023building } propose prompt tuning for adapting language models. In the visual domain, ~\cite{jia_visual_2022} introduces visual prompt tuning as a highly efficient and effective approach for large-scale Transformer models in vision. Another efficient fine-tuning alternative is bias-tuning~\cite{cai_tinytl_2020, zaken2021bitfit} which is a sparse fine-tuning method that only fine-tunes a subset of the bias terms of the model during training.
\paragraph{Federated Learning}
Federated learning is an innovative machine learning technique enabling model training across numerous decentralized devices or servers, all while avoiding the need to transfer data to a central server~\cite{mcmahan2017communication, zhao2018-non-iid}. Until now, FedAvg~\cite{mcmahan2017communication} has been widely considered the primary benchmark for federated learning. In FedAvg, the weight \textit{aggregation} is performed by averaging the model weights obtained from different devices. FedAvg works well when the data is homogenous. However, in heterogeneous environments, the global model that is suitable for all clients faces convergence challenges. FedProx\cite{li2020federated}, FedAdapt \cite{li2020fedadapt}, \cite{zhao2018federated}, FedNova~\cite{FedNova-2020} and SCAFFOLD~\cite{karimireddy2020scaffold} are some enhancements proposed to FedAvg which have attempted to develop modified versions of the algorithm that can handle non-IID data. In our evaluation, we establish that FedAvg has the tendency to achieve satisfactory performance in the heterogeneous environment given a pre-trained foundation model. 

Although there are some recent works on FL for HRI~\cite{GAMBOAMONTERO2023118510, 9983628}, no study has yet explored HRI for action recognition. Our research constitutes the initial investigation into the video understanding of HRI within Federated Learning (FL) settings. Additionally, this study serves as a fundamental basis for the utilization of parametric-efficient fine-tuning in conjunction with pre-trained models for the purpose of video comprehension within FL.
\section{Efficient Federated Learning for Human Robot Interaction Recognition}
\subsection{Federated Learning in HRI Action Recognition}\label{subsec:CFL}
Industrial HRI action recognition is a multi-class classification task where, given $T$ training samples of a video dataset $X^T=\{(x_t,y_t)_{t=1}^T\}$, with $y \in \{0,1,..,C-1\}$ for $C$ classes, the goal is to achieve accurate classification on a set of unseen videos $X^u=\{x_1,...,x_M\}$, where $M$ is the number of videos in the test set. To set up the HRI recognition task in the FL setting, we assume a system of $N$ clients that can coordinate with the centralized server without sharing their local data. Further, let $\mathcal{X}$ be a subset of $\mathbb{R}^p$ representing the instance space, $\mathcal{Z}$ a subset of $\mathbb{R}^d$ denoting the latent feature space and $\mathcal{Y}$ a subset of $\mathbb{R}$ representing the output space. The server model $F$, parameterized by $\Theta := [\vtheta^f; \vtheta^p]$, comprises two components: a feature extractor $f: \mathcal{X} \to \mathcal{Z}$ parameterized by $\vtheta^f$, and a predictor $p: \mathcal{Z} \to \Delta^\mathcal{Y}$ parameterized by $\vtheta^p$, where $\Delta^\mathcal{Y}$ is the simplex over $\mathcal{Y}$.

In each communication round, the server randomly selects a subset of available clients $\sS_r \in [\sN]$ with $|\sS_r| = n$ and broadcasts the model to all clients. The clients, upon receiving the model from the server, perform several steps of stochastic gradient descent (SGD) updates on their local training data $T_k \subset X^T$, obtain the updated model, and send their model parameters $\Theta := [\vtheta^f_k; \vtheta^p_k]_{k=1}^{n}$ back to the server.

Assuming the global model is initialized by $F(\Theta)$, the clients minimize their loss in each round $r$ using SGD training as follows:
\begin{equation}
    \min_{\Theta} F^{(r)}_{k}(\Theta) = \frac{1}{|T_k|} \sum_{t=1}^{|T_k|} \ell_k(\Theta,x_t),
\end{equation}
\noindent  where $\ell$ is the loss function, $r$ is the communication round, and $|T_k|$ represents the number of the training samples of the $k$th client. The server collects all the parameter updates from clients and conducts model averaging as~\cite{mcmahan2017communication}
 \begin{equation}
\Theta^{(r+1)} = \sum_{k=1}^n \frac{|T_k|}{\sum_{i=1}^n|T_i|}\Theta_k^{(r)}.
 \label{eq:client}
 \end{equation}
This parameter exchange between the clients and server goes on from $r=0$ to $r=R-1$ until the convergence of the global model.
\subsection{Problem definition}
As stated in Section~\ref{subsec:CFL}, the global model convergence in FL is accomplished after several rounds of parameter exchange between the local nodes and the global server. The procedure is repeated for $r$ rounds. In this paper, we let $\sC$ represent the communication cost associated with the FedAvg baseline. This cost is directly related to the number of parameters shared by the clients as $ \sC\propto |\vtheta| \cdot |\sS_r|$, where $ \vtheta \subseteq \Theta$  the parameters that need to be transmitted. 
% with the server as shown below
% In this paper, we compute the communication cost for FedAvg~\cite{mcmahan2017communication}  baseline $\sC$ which is directly proportional to the number of parameters shared by the clients to the server as 
\iffalse
\begin{equation}
    \sC\propto |\vtheta| \cdot |\sS|,
\end{equation}
% \SV{something is missing here. There is no F and $F_k$ as described below}
% $F$ is the global model, $F_k$ is the client model,  
%and $\vtheta << \Theta$,  represent
\noindent where $ \vtheta \subseteq \Theta$  the  parameters that need to be transmitted. 
\fi 
We aim to minimize $\sC$ without an accuracy drop for the  HRI recognition task.
As the communication cost is directly related to the number of trainable parameters, one intuitive method is to freeze the backbone $\vtheta^f$ and only train the penultimate linear layer $\vtheta^p$. However, the drawbacks of linear probing in terms of performance, which will be discussed in Section~\ref{results}, have prompted us to introduce our federated learning framework for human-robot interaction recognition \textbf{(CEFHRI)} which keeps tunable parameters, $\vtheta << \Theta$ using Vision Tansformers~\cite{dosovitskiy_image_2021}.
\subsection{Proposed Framework}
In this section, we provide a succinct overview of the Vision Transformer (ViT) and its application in the video understanding task of FL followed by an introduction to spatial-temporal adaptation. We further discuss the utility of such adaptation to preserve server privacy.
\paragraph{Overview}
The transformer architecture~\cite{dosovitskiy_image_2021} has been extensively utilized in vision applications such as video surveillance and action recognition. In this work, we investigate the role of FL in recognizing human-robot interaction which is another video understanding task. To this end, we introduce CEFHRI, an FL framework that employs the vision transformer for efficient decentralized HRI learning.  Within the CEFHRI framework, the local and global models are customized by inserting an adapter module.  Taking inspiration from the adapter design in~\cite{chen_adaptformer_2022, pan_parameter-efficient_2022}, we propose a variant of spatio-temporal adapter~\cite{pan_parameter-efficient_2022} architecture, called the ST-Adapter specifically designed for a video pre-trained model. 

\begin{figure}[t!]
\centering
 \includegraphics[scale = 0.25]{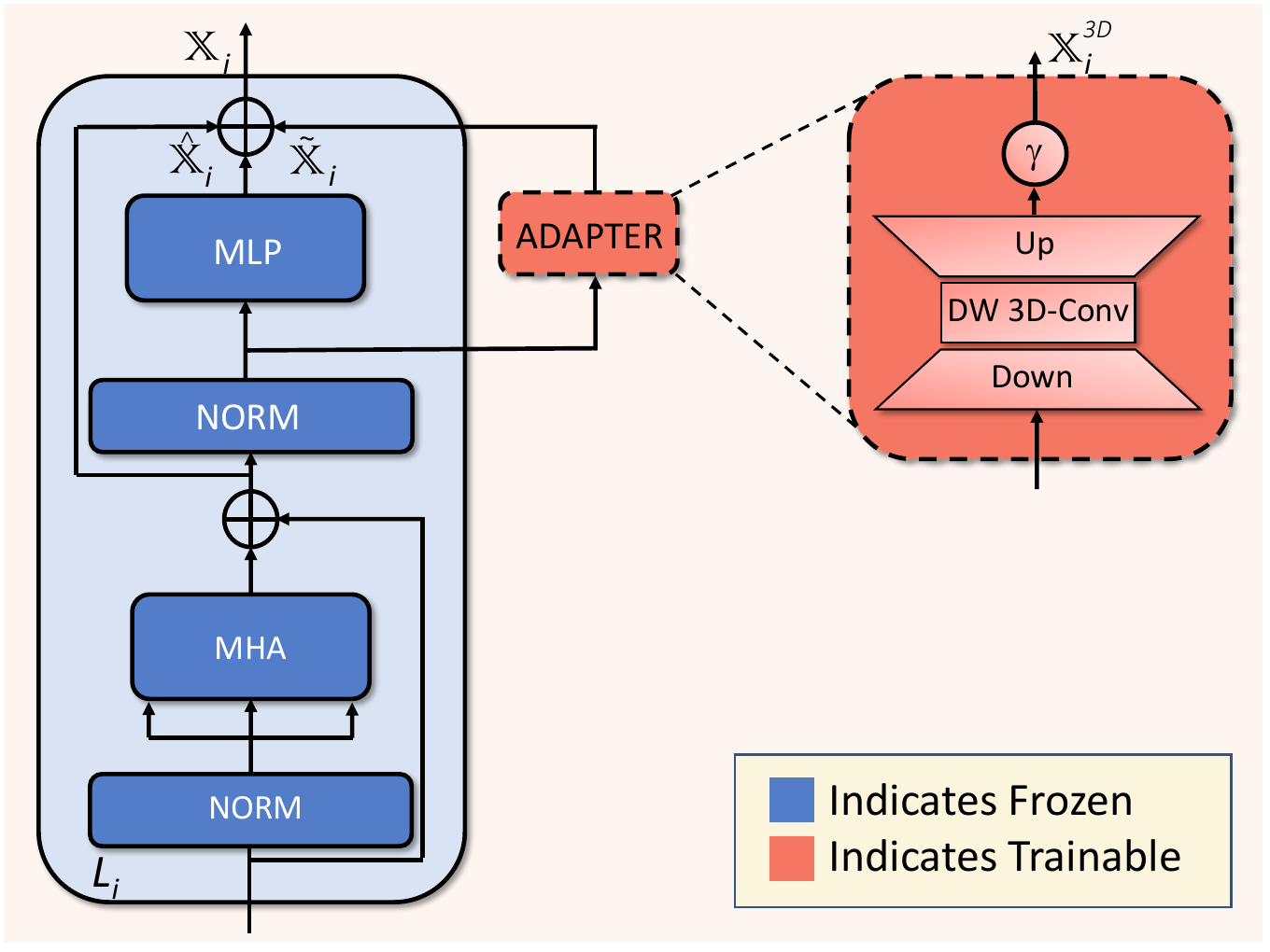}
\caption{\footnotesize The adapter design of the proposed CEFHRI framework. The adapter is inserted in the $i^{th}$ layer of the backbone model while keeping other blocks frozen.}
\label{fig:adapter}
\end{figure}
\begin{figure}[b!]

\centering
 \includegraphics[scale = 0.5]{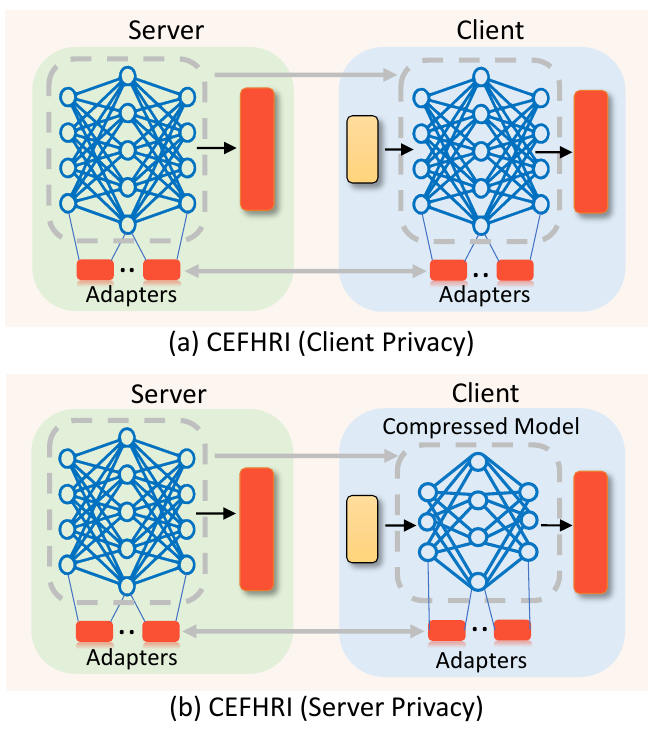}
\caption{\footnotesize (a) CEFHRI with client data privacy. (b) CEFHRI with server model privacy }
\label{fig:server_privacy}
\end{figure}
\paragraph{Adapter Design} 
% The designed adapter architecture preserves the spatial and temporal information of videos while enabling efficient fine-tuning.
The designed adapter architecture allows for the preservation of both the spatial and temporal characteristics of videos, while simultaneously facilitating efficient fine-tuning. The present discourse will concentrate on creating adapters for deep transformer backbones. In particular, we have introduced a depth-wise 3D convolution layer in a standard configuration, placed between the bottlenecks of the vanilla adapter~\cite{chen_adaptformer_2022} and parallel to the MLP block. The visual representation of this arrangement is depicted in the accompanying Fig.~\ref{fig:adapter}. 

% Specifically, we introduce a standard depth-wise 3D convolution layer between the bottlenecks of the vanilla adapter~\cite{chen_adaptformer_2022} parallel to MLP block, as shown in Fig.~\ref{fig:adapter}. 

Since the ST-Adapter branch is inserted alongside the MLP block, we term the adapted MLP block as the ST-adapted MLP and its output as $\sX^{3D}$. The 3D branch will transform the features in the following manner:
\begin{equation}
\Tilde{\sX}_{i}=\gamma \cdot ((DWConv3D(NL(\hat{\sX}_{i})\cdot\ W_{DOWN}))\cdot W_{UP})
\end{equation}
where $DWconv3D$ represents the depth-wise 3D-convolution. Here, $\gamma$ is a tuneable parameter to scale the adapter output as in \cite{chen_adaptformer_2022}. Before applying $DWConv3D$, the features are reshaped from $\sX_{i}^{'}\in \sR^{N\times \hat{d}}$ to $\sX_{i}^{''}\in \sR^{D\times H \times W \times \hat{d}}$, where $\sX^{'}= NL(\hat{\sX}_{i})$. $\Tilde{\sX}^{i}$ features are then fused with the MLP branch output features to generate $\sX^{3D}$, 

\begin{equation} \sX_{i}^{3D}=\mathbf{MLP}(NL(\hat{\sX}_{i}))+\Tilde{\sX}_{i}\vspace{-1mm}
  \end{equation}
Further, $\hat{\sX}$ features are fused with $\sX^{3D}$ using a residual connection as following,
  \begin{equation}
      \sX_{i}=\sX_{i}^{3D} + \hat{\sX}_i
  \end{equation}
We provide a detailed comparison of the proposed adapter design with other parametric-efficient finetuning techniques in Section~\ref{results} under various FL settings. We further evaluate the existing adapter designs from the literature in Section~\ref{ablations}.
 \begin{algorithm}
\caption{The CEFHRI Framework}\label{alg:CEFHRI}
\small
\begin{algorithmic}
\item \hspace{-3mm}
\noindent \colorbox[rgb]{1, 0.95, 1}{
\begin{minipage}{0.9\columnwidth}
\textbf{Server:} initialize the foundation model $F$ parametrized by $ \Theta$, and the compressed model $F^c$  with~$ \Theta^c$\\
\textbf{Clients:} fetch the compressed model weights $ \Theta^c$, and initialize the tunable adapter weights, $\vtheta$\\
\textbf{Require:} $\sN$ clients, sampling rate $\sR\in(0,1]$, communication round budget $R$, shared weight $\vtheta \subseteq \Theta$.
\end{minipage}
}
\item \hspace{-3mm}
\colorbox[rgb]{1, 0.97, 0.91}{
\begin{minipage}{0.915\columnwidth}

\item      \textbf{for } each round $r=0,1,2,...,R-1$ \textbf{do}
% \item     \hspace*{\algorithmicindent}  $n \leftarrow {\rm{max}}(\sR \cdot \sN,1)$
\item     \hspace*{\algorithmicindent}  $\sS_r\leftarrow$ Server samples $|\sS|$ clients from $N$ clients
\end{minipage}}
\item \hspace{-3mm}
\colorbox[rgb]{0.95, 0.98, 1}{
\begin{minipage}{0.9\columnwidth}
\item ~~ \textbf{Client update ($k, \Theta^{c}, \vtheta^{(r)}$):}
\item   ~~  \hspace*{\algorithmicindent}  \textbf{for} each client $k\in\sS_r$ \textbf{do}
% \item     \hspace*{\algorithmicindent} $F_k \leftarrow$ load $\sP$ with $\vtheta$
\item   ~~  \hspace*{\algorithmicindent} \quad $F_k \leftarrow$ \{$\Theta^{c},\vtheta_k^{(r)}$\}
\item  ~~   \hspace*{\algorithmicindent} \quad training $F_k$ by SGD on $\sD_k$ for E epochs
\item ~~ \hspace*{\algorithmicindent} \quad $\vtheta^{(r+1)}_k \leftarrow \vtheta^{(r)}_k-\eta \nabla F_k $
% \item     \hspace*{\algorithmicindent} $\vtheta\leftarrow$ weights for parameters in $\sP$ of $F^{(r)}_k$

% \item     \hspace*{\algorithmicindent} $\vtheta\leftarrow$ weights for parameters in $\sP$ of $F^{(r)}_k$
\item   ~~  \hspace*{\algorithmicindent} \quad return $\vtheta^{(r+1)}_k$ to server
\end{minipage}
}

\item \hspace{-3mm}
\colorbox[gray]{0.95}{
\begin{minipage}{0.9\columnwidth}
\item ~~ \textbf{Server executes:}

\item   ~~  \hspace*{\algorithmicindent}  \textbf{for} each client $k\in\sS_r$ \textbf{do}
% \item     \hspace*{\algorithmicindent}  \quad $\vtheta^{(r)}\leftarrow$ weights for parameters in $\sP$ of $F^{(r)}$
% \item \hspace*{\algorithmicindent}  \quad  download $\Theta ^{(r)}$ from the server
\item   ~~  \hspace*{\algorithmicindent}  \quad $\vtheta_k^{(r+1)} \leftarrow \text{ClientUpdate}(k,\Theta^{c}, \vtheta^{(r)})$
% \item     \hspace*{\algorithmicindent} \quad \quad $f_m^{(t)\leftarrow}$ load $\sP$ with $\vtheta^{(t)}$
 \item  ~~~~~   \hspace*{\algorithmicindent} $ \theta^{(r+1)}\leftarrow$ aggregate updated parameters $\vtheta_k^{(r+1)}$~by
 \item \hspace{2.3cm} clients as in Eq.~\ref{eq:client}
\end{minipage}
}
\item \hspace{-3mm}

\end{algorithmic}
\end{algorithm}

\begin{table*}[t]
  \caption{\footnotesize Human-robot interaction action recognition performance in terms of \% accuracy under FL settings where maximum communication rounds, $R=40$. All results represent the HRI recognition accuracy for different FL settings across three datasets. Here, $N$ and $|\sS_r|$ indicate total clients and sampled clients respectively.}
		\centering
		\scalebox{0.85}{
			{
				\begin{tabular}{l|c|c|c|c|c|c|c|c|c|c}
					\hline
	
					\multirow{2}{*}{\textbf{Clients}}             & \multirow{2}{*}{\bf Method}  &              \multicolumn{3}{c|}{\bf COIN }               &                     \multicolumn{3}{c|}{\bf InHARD}                     &                   \multicolumn{3}{c}{\bf HRI30}                                                      \\
					&                          & $ \alpha = 0.1$  & $\alpha = 0.5$ & $\alpha = 1.0$&   $ \alpha = 0.1$  & $\alpha = 0.5$ & $\alpha = 1.0$&  $ \alpha = 0.1$  & $\alpha = 0.5$ & $\alpha = 1.0$      \\ \hline
					
					\multirow{6}{*}{$N=16, |\sS_r|=16$}  &           Full Fine-Tuning   &46.1\small$\pm{0.1}$&46.9\small$\pm{0.1}$&47.3\small$\pm{0.2}$&81.3\small$\pm{0.2}$   &82.4\small$\pm{0.1}$&83.2\small$\pm{0.1}$  &76.2\small$\pm{0.2}$   &81.5\small$\pm{0.1}$&87.2\small$\pm{0.1}$              \\
					\cline{2-11}
					&     Linear Probing&33.4\small$\pm{0.1}$ &34.1\small$\pm{0.1}$ &34.4\small$\pm{0.2}$ &44.7\small$\pm{0.2}$ &51.2\small $\pm{0.4}$&55.6\small$\pm{0.1}$  & 35.1\small$\pm{0.1}$&36.6\small$\pm{0.1}$&42.6\small$\pm{0.2}$       \\
					%\cline{2-10}
					&Bias-Tuning&33.7\small$\pm{0.1}$&34.6$\pm{0.1}$&34.8\small$\pm{0.1}$&47.5\small$\pm{0.2}$&54.2\small$\pm{0.1}$&60.6\small$\pm{0.1}$& 45.7\small$\pm{0.1}$&48.8\small$\pm{0.1}$&50.1\small$\pm{0.2}$\\
					&Prompt-Tuning&36.1\small$\pm{0.1}$&36.9\small$\pm{0.1}$&37.1\small$\pm{0.2}$&69.8\small$\pm{0.1}$&70.6\small$\pm{0.3}$&71.1\small$\pm{0.1}$ &67.6\small$\pm{0.1}$&72.4\small$\pm{0.1}$&75.8\small$\pm{0.2}$\\
					&CEFHRI&\textbf{44.6}\small$\pm{0.1}$&\textbf{45.1}\small$\pm{0.1}$&\textbf{45.8}\small$\pm{0.3}$&\textbf{80.6}\small$\pm{0.1}$&\textbf{81.3}\small$\pm{0.1}$&\textbf{82.1}\small$\pm{0.3}$ &\textbf{73.6}\small$\pm{0.1}$&\textbf{81.2}\small$\pm{0.1}$&\textbf{85.8}\small$\pm{0.2}$\\
					%\cline{2-10}	
					 \hline
			\multirow{6}{*}{$N=16, |\sS_r|=4$}  &            Full Fine-Tuning&38.8\small$\pm{0.3}$&43.9\small$\pm{0.1}$&45.2\small$\pm{0.3}$ &73.56\small$\pm{0.2}$ &74.4\small$\pm{0.1}$&77.2\small$\pm{0.1}$  & 73.8\small$\pm{0.1}$&82.6\small$\pm{0.1}$&85.1\small$\pm{0.2}$                 \\
			\cline{2-11}
					&     Linear Probing&29.4\small$\pm{0.3}$&32.6\small$\pm{0.2}$&32.9\small$\pm{0.3}$&33.3 \small$\pm{0.2}$&38.8\small$\pm{0.1}$ & 42.6  \small$\pm{0.1}$& 39.6\small$\pm{0.1}$&35.6\small$\pm{0.1}$&34.8\small$\pm{0.2}$     \\
					%\cline{2-10}
					& Bias-Tuning&30.5\small$\pm{0.2}$&32.9\small$\pm{0.2}$&33.4\small$\pm{0.3}$&39.5\small$\pm{0.2}$&42.2\small$\pm{0.1}$&48.1\small$\pm{0.1}$& 49.3\small$\pm{0.1}$&47.6\small$\pm{0.1}$&45.0\small$\pm{0.2}$\\   
					&Prompt-Tuning&32.4\small$\pm{0.2}$&35.2\small$\pm{0.2}$&36.5\small$\pm{0.2}$ &63.1\small$\pm{0.1}$&65.4\small$\pm{0.1}$&68.8\small$\pm{0.4}$&64.6\small$\pm{0.1}$&72.6\small$\pm{0.1}$&74.7\small$\pm{0.2}$\\
					%\cline{2-10}
					& CEFHRI&\textbf{38.6}\small$\pm{0.1}$&\textbf{43.1}\small$\pm{0.1}$&\textbf{44.2}\small$\pm{0.2}$&\textbf{72.9}\small$\pm{0.2}$&\textbf{73.3}\small$\pm{0.3}$&\textbf{76.5}\small$\pm{0.1}$&\textbf{73.2}\small$\pm{0.1}$&\textbf{80.6}\small$\pm{0.1}$&\textbf{81.5}\small$\pm{0.2}$
			       \\ \hline
		\end{tabular}}}
		\label{tab:table1}
	\end{table*}
 
\paragraph{Server-Side Privacy}
We have additionally expanded the applicability of the proposed adapter design to a situation wherein the foundation model owner, i.e., the server is precluded from disclosing their model to the data owner~\cite{xiao2023offsite}. To overcome this challenge, we devise a strategy to fine-tune the proposed adapter on the data owner's dataset without accessing the complete pre-trained model weights. In order to safeguard model ownership while simultaneously enhancing efficiency, we have implemented a technique known as lossy compression on the frozen backbone model as shown in Fig.~\ref{fig:server_privacy}. We leverage the findings of~\cite{naseer2021intriguing} which reveal that the discriminative information crucial for accurate classification is predominantly captured within the class tokens of the final few blocks. Specifically, we have selectively dropped a few layers from the model to produce a compressed variant, an adapter counselor, $F^c$  with parameters $\Theta^c$ such that  $\Theta^c< \Theta$.
The purpose of utilizing $F^c$ is to furnish approximate gradient directions to update the adapters, while simultaneously maintaining similarity to the original frozen component weights, $\Theta$. Nonetheless, it is imperative that the $F^c$ precision be restrained, as a higher degree of accuracy could potentially divulge information regarding the original model. Furthermore, a smaller $F^c$ size facilitates a more efficient fine-tuning process for downstream users. 
We report our results for various compression ratios in Section~\ref{ablations}. The outline of the proposed framework with client data privacy and server model privacy has been described in Algorithm~\ref{alg:CEFHRI}.
\section{Experiments}
We evaluate CEFHRI for downstream human-robot interaction tasks across a wide range of FL settings. The experimental setup is described in Section~\ref{experiment_setup}, and the effectiveness of CEFHRI is demonstrated in Section~\ref{results}.
\subsection{Experiment Setup}\label{experiment_setup}
% \textbf{Datasets:} 
%  The InHARD dataset contains 13 different industrial action classes and over 4800 action samples. The HRI30 dataset contains 30 categories of industrial-like actions and 2940 manually annotated clips. we used 10337 videos distributed over 180 classes in COIN Dataset. 
\paragraph{Architecture and Datasets} In our experiments, we use Kinetics-400~\cite{kay2017kinetics} dataset for pre-training. For the downstream FL task, we select three datasets with varying degrees of domain gap as compared to Kinetics-400:~$(i)$ InHARD~\cite{dallel2020InHARD}, $(ii)$~HRI30~\cite{iodice2022hri30} and $(iii)$~COIN Dataset~\cite{tang2019COIN}. We adopted experimental settings from~\cite{chen_adaptformer_2022, tong2022videomae}. We use plain ViT-B/16~\cite{dosovitskiy_image_2021} with supervised pre-train weights from VideoMAE~\cite{tong2022videomae} official repository, where pre-trained checkpoints on Kinetics-400~\cite{kay2017kinetics} dataset are publicly available. An extra BatchNorm layer~\cite{ioffe2015batch} without affine transformation is inserted before the penultimate layer as in~\cite{chen_adaptformer_2022}. For all datasets, we use 8 frames with a temporal stride of 4. The tubelet size is set to 2 as in VideoMAE's default settings ~\cite{tong2022videomae}. The VideoMAE codebase includes a downsampling layer that converts original frames based on the tubelet size ratio. Hence, the final number of tokens for the transformer block is $4\times14\times14$, where 4 is the downsampled number of frames while the original input video has 8 frames.
% The InHARD dataset~\cite{dallel2020InHARD} dataset comprises 13 distinct industrial action classes, encompassing more than 4800 action samples. The HRI30~\cite{iodice2022hri30} database is a collection of videos specifically curated for industrial action recognition. It encompasses 30 categories of industrial-like actions and consists of 2940 clips. The COIN dataset~\cite{tang2019COIN}, designed for comprehensive instructional video analysis, is a vast collection comprising 11,827 videos. These videos cover 180 distinct tasks across 12 domains, encompassing various aspects of our daily lives such as vehicles, gadgets, and more.
% We have ignored the \textbf{[CLS]} token in our implementation based on ~\cite{dosovitskiy_image_2021} which establishes that [CLS] token has no significant effect on the vision transformer performance.
\paragraph{Evaluation Metrics} 
We compare the proposed CEFHRI  with four commonly used fine-tuning baselines: (1) \emph{Full Fine-tuning}, (2) \emph{Linear Probing (fine-tuning classification head only)}, (3) \emph{Bias-Tuning}, and (4) \emph{Prompt-Tuning}. Here full fine-tuning indicates that all the parameters of the foundation model are fine-tuned on the downstream dataset. Linear probing indicates that only the penultimate linear layer is fine-tuned. Bias-tuning~\cite{cai_tinytl_2020, Bitfit-2022} aims to fine-tune only the bias parameters for the downstream task while prompt-tuning~\cite{prompt2021,jia_visual_2022} concatenates prompt tokens to the input embedding, where each prompt token is a learnable $d$-dimensional vector. The baselines are compared with CEFHRI with respect to two evaluation criteria \textbf{$(i)$ Communication Efficiency, $(ii)$ Human-Robot Interaction recognition}.

\paragraph{Training Settings}
We distribute the training data between clients based on the Dirichlet distribution (with $\alpha \in \{0.1, 0.5, 1.0\}$) to achieve heterogeneous data partitioning across clients. Lower $\alpha$ indicates a higher degree of data heterogeneity~\cite{curr-vahid}. In all experiments, we assume $N=16$ clients are available, and we set the sampling rate to $25\%$, and $100\%$ which means either $|\sS_r|=4$ or $|\sS_r|=16$ in each round. Each client performs $E=8$ local epochs with a batch size of $8$ before global aggregation is performed for  $R=40$ communication rounds. We keep the model weights frozen during the fine-tuning of CEFHRI techniques. We set $\gamma = 2.5$, while the bottleneck dimension, $\hat{d}=64$ in our default settings for the adapter. We followed the training settings of~\cite{chen2022adaptformer} except the learning rate which we selected as 0.001. For prompt-tuning, the number of introduced tokens is set to 8 based on optimal performance as mentioned in~\cite{chen_adaptformer_2022}. Any modification to the default settings will be clearly stated.
\begin{table*}[t]
\caption{\footnotesize The \textbf {communication cost} is computed with 4Bytes/parameter with $N=\sS=16$, $\alpha = 0.5$, and $R=40$. The number in the bracket indicates the target action recognition accuracy \%. \textemdash    ~indicates that the target accuracy is not achieved under that particular setting. \textbf{Parameters} are the trainable parameters of each client. $\downarrow$ indicates smaller values are better.}	
		\centering
		\scalebox{0.75}{
			{
				\begin{tabular}{l|c|c|c|c|c|c|c|c|c}
					\hline
	
					 \multirow{2}{*}{\bf Method}  &              \multicolumn{3}{c|}{\bf COIN (45)}               &                     \multicolumn{3}{c|}{\bf InHARD (80)}                     &                   \multicolumn{3}{c}{\bf HRI30 (80)}                                                      \\
	& \bf Parameters (M)$\downarrow$ & \bf Rounds$\downarrow$& \bf Comm. cost$\downarrow$  & \bf Parameters (M) $\downarrow$ & \bf Rounds $\downarrow$  & \bf Comm. cost $\downarrow$   &  \bf Parameters (M)$\downarrow$& \bf Rounds $\downarrow$&\bf Comm. cost $\downarrow$   \\ \hline
					
					       Full Fine-Tuning &86.36&20&5.39$\times$ 20 GB&    86.23 &14&5.38$\times$14 GB  &86.25&21&5.39$\times$21 GB              \\
					       \hline
					  Linear Probing &0.13&-&-&0.01&-&-&0.01&-&-    \\
					%\cline{1-9}
					Bias-Tuning&0.23&-&-&0.10&-&-&0.10&-&-\\
					Prompt-Tuning&0.20&-&-&0.08&-&-&0.09&-&- \\
					CEFHRI&1.33&35&85.12$\times$ 35 MB&1.20&38&76.80$\times$38 MB&1.21&37&77.44$\times$37 MB\\					
					 \hline		
		\end{tabular}}}
		\label{tab:table2}
  %\vspace{-4mm}
	\end{table*}
\subsection{Main Results}\label{results}
The results reported in this section ensure the client's data privacy with the assumption that the client has access to the non-compressed version of the pre-trained model.

\begin{figure}[h!]
 \includegraphics[scale=0.35]{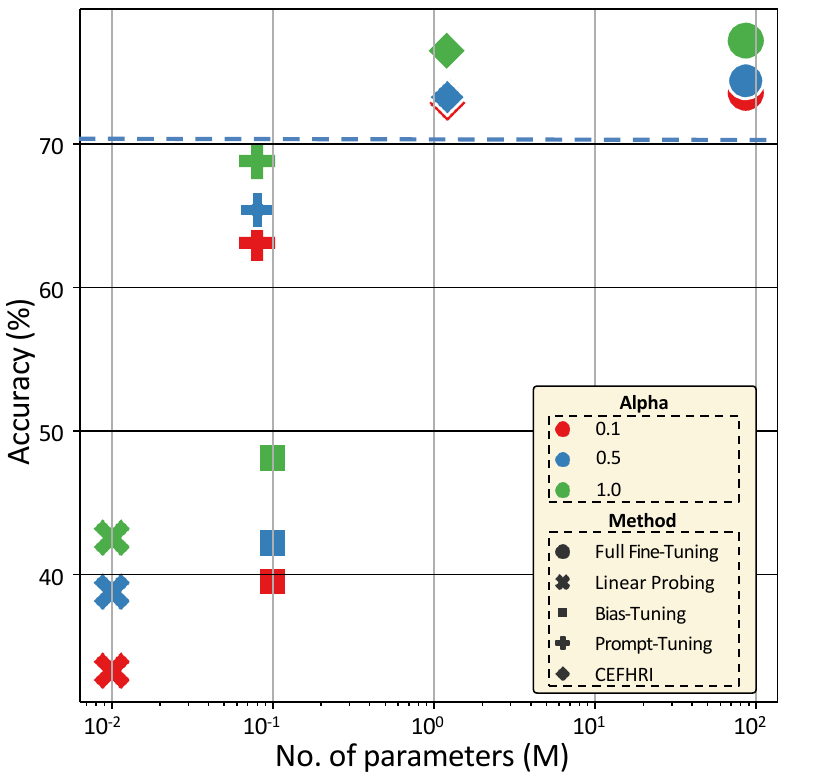}
 \centering
 %\vspace{1mm}
\caption{\footnotesize Federated training accuracy vs \# of parameters plot for all fine-tuning methods under consideration.  The figure illustrates that only CEFHRI can achieve the target accuracy of 70\% on InHard dataset given $R=40$, $N=16$, and $|\sS_r|=4$ for numerous levels of data heterogeneity. }
\label{fig:accvsparam}

\end{figure}

\paragraph{Human-Robot Interaction Recognition} Table~\ref{tab:table1} displays the interaction recognition results of the CEFHRI framework. Our assessment shows that the CEFHRI framework surpasses all other approaches, except for full fine-tuning, across all datasets and FL configurations. As reported in Table~\ref{tab:table1}, the CEFHRI framework achieves satisfactory adaptation performance despite having significantly fewer trainable parameters than full fine-tuning. Moreover, the results presented in Table~\ref{tab:table1} indicate that irrespective of the value of $\alpha$, there is no substantial difference in the performance gap for CEFHRI. This highlights the crucial role of the pre-trained model in mitigating the impact of heterogeneous data. These findings are in agreement with recent studies conducted on image classification tasks~\cite{chen_pre-training_2022,nguyen_where_2022}.
\paragraph{Communication Cost}
In this paper, we only consider the uploading \emph{(i.e., from clients to server)} communication cost, $\sC$, for FedAvg~\cite{mcmahan2017communication} baseline defined as $\sC=R\times |\vtheta| \times |\sS_r| \times$ 4B, where $R$ stands for the total number of communication rounds; $|\sS_r|$  denotes the number of participating clients, and $ |\vtheta| \subseteq |\Theta|$  represents the subset of the total number of model parameters\footnote{Each dataset has different \# of classes, so the \# of model parameters are different owing to varying linear classification head parameters.} that are exchanged between each client and server in each round, with each parameter occupying 4 bytes (4B) of storage. With this in mind, we report the communication efficiency results against the pre-specified target accuracy for each dataset in Table~\ref{tab:table2}. Upon comparison to full fine-tuning, it becomes evident that the CEFHRI framework can achieve the target accuracy with a significantly reduced communication cost of at least $\approx$ 35$\times$ lower. Despite the \emph{linear probing} approach having the fewest tunable parameters, it falls short of attaining the target accuracy for any of the datasets being examined. Fig.~\ref{fig:accvsparam} further illustrates the performance vs cost tradeoff between the baselines and CEFHRI. This suggests that the proposed parameter-efficient fine-tuning prototype of the CEFHRI framework is efficacious in mitigating communication overhead, without compromising recognition performance.
\paragraph{CEFHRI vs Parametric Efficient Model}One straightforward solution to reduce the communication cost is to use tiny video networks such as the X3D-S model \cite{feichtenhofer2020x3d}. Nevertheless, the architectures with low capacity are incapable of achieving satisfactory performance even with pre-training, as is evidenced in Table~\ref{tab:table3}. In particular, X3D-S \cite{feichtenhofer2020x3d} could only achieve 30.8\% HRI recognition performance on HRI30~\cite{iodice2022hri30}, while CEFHRI yields 84.6\% accuracy for $\alpha=0.5$. Our comprehension is that models with low capacity are not effective in producing desirable performance under data heterogeneity scenarios. Consequently, the CEFHRI approach offers a practical remedy for mitigating the decrease in performance while still keeping the communication cost as minimal as that of smaller models, such as X3D-S, for the purpose of HRI recognition.
\begin{table}[b!]
\caption{\footnotesize {Performance comparison with parameter-efficient model \textbf{X3D-S}. Here, $R=40$ , $N=16$, and $|\sS_r|=4$. All results represent the averaged percentage accuracy of HRI recognition over several runs for $\alpha = 0.5$. $\uparrow$ indicates larger values are better and $\downarrow$ indicates smaller values are better. \# of parameters indicates the tunable parameters.}}			
		\centering
 		\resizebox{0.5\textwidth}{!}{
			{
				\begin{tabular}{l|c|cc}
					\hline
						\textbf{Dataset}&\textbf{\bf Architecture }       &  \textbf{\# of Parameters $\downarrow$}       &    \bf Accuracy(\%) $\uparrow$                                                                                   \\
				 \hline
					
				\multirow{2}{*}	{COIN}&  X3D-S &           3.34M	&15.9\small$\pm{0.3} $                 \\
					
					 %\cline{2-5}
			 &CEFHRI  &    1.33M& 43.1\small$\pm{0.1}$                \\
					 \hline
			       \multirow{2}{*}	{InHARD}&  X3D-S &         3.00M &26.9\small$\pm{0.1}     $         \\

					 %\cline{2-5}
			&CEFHRI  &           1.20M&        73.3\small$\pm{0.3}$            
			       \\ \hline
			       \multirow{2}{*}	{HRI30}&  X3D-S &          3.03M&30.8\small$\pm{0.1}$             \\
					
					 %\cline{2-5}
			&CEFHRI&1.21M&         80.6\small$\pm{0.1}$             \\
					\hline
		\end{tabular}}
		}
		\label{tab:table3}
% \end{wraptable}
\end{table}

\section{Ablation Studies}
\label{ablations}
\paragraph{Impact of Model Initialization}
To demonstrate the applicability of the pre-trained model in the CEFHRI framework, we undertake a more in-depth examination of two fundamental questions:~\textbf{(1)} How does the model initialization impact the industrial HRI action recognition performance in FL settings? \textbf{(2)} To what extent can pre-training alleviate the accuracy drop caused by the heterogeneity of data from clients in the context of industrial HRI action recognition?

\begin{figure}[htbp!]
 \includegraphics[scale=0.28]{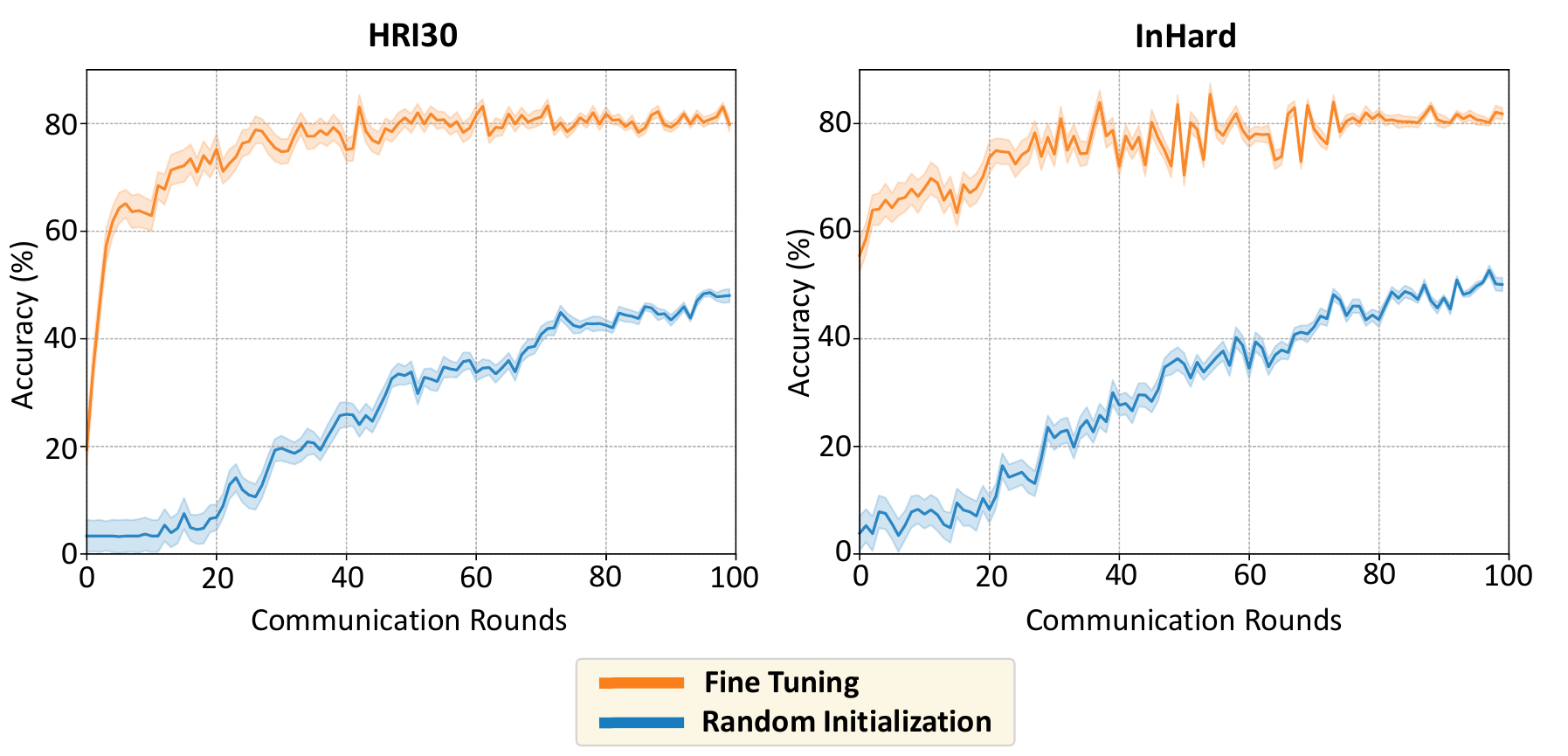}
 \centering
\caption{\footnotesize FL results for the industrial HRI action recognition task reveal the advantages of utilizing the pre-trained model. The figure illustrates that employing a pre-trained model in FL saves communication costs while achieving superior performance. Here, $R=100$, $N=16$, and $|\sS|=16$.}
\label{fig:plots}
\end{figure}

Our findings show that using a pre-trained model can drastically improve the performance of industrial HRI action recognition in FL. Fig.~\ref{fig:plots} demonstrates the significant advantage of using a pre-trained model over training from scratch for the InHARD~\cite{dallel2020InHARD}, and ~HRI30~\cite{iodice2022hri30} datasets. It is evident that the server achieves notable performance within a fewer number of rounds. This is in contrast to random initialization, which is unable to narrow the performance gap even after 100 rounds for heterogeneous FL settings. \emph{Our results indicate that using pre-trained models not only outperforms random initialization but also effectively mitigates the data heterogeneity effect in the industrial HRI action recognition task.} 
\paragraph{Performance Gain of the Proposed Adapter Design}
\label{adapter_comparison}
CEFHRI adapter is different in design from ~\cite{chen2022adaptformer} and distinct in application from~\cite{pan_parameter-efficient_2022}. Firstly, the ST-adapter proposed by~\cite{pan_parameter-efficient_2022} is specifically designed for frozen CLIP models, and its effectiveness in FL has not been explored. Secondly, the sequential implementation doesn't maintain the original features achieved by the frozen backbone. We argue that the parallel design of the adapter maintains the original features as the adapter branch is added parallel to the original MLP block and only scaled feature aggregation is performed. As shown in Table~\ref{tab:table5}, our analysis indicates that the parallel adapter structure outperforms the sequential design. As a result, we use the parallel ST-adapter design by default for the CEFHRI prototype, which distinguishes it structurally from that of~\cite{chen_adaptformer_2022}. 
\begin{table}[hbt!]
\caption{\footnotesize {Performance comparison with different adapter designs. Here, $R=40$ , $N=16$, and $|\sS_r|=4$. All results represent the averaged percentage accuracy of HRI action recognition over several runs for $\alpha = 0.5$.}}			
		\centering
  \scalebox{0.70}{
 		\resizebox{0.5\textwidth}{!}{
			{
				\begin{tabular}{l|c|c}
					\hline
				\textbf{Dataset}&\textbf{\bf Adapter Design }              &    \bf Accuracy(\%)                                                                                 \\
				 \hline					
				% \multirow{3}{*}	{COIN}&  Adaptformer~\cite{chen_adaptformer_2022} &  15.9 \small$\pm{0.3} $ \\
				% 	&CLIP-Adapter~\cite{pan_parameter-efficient_2022}  &     43.1\small$\pm{0.1}$\\
				% 	 %\cline{2-5}
			 %         &CEFHRI-Adapter  &  43.1\small$\pm{0.1}$                \\
				% 	 \hline
			       \multirow{3}{*}	{InHARD}&  Adaptformer~\cite{chen_adaptformer_2022} &  55.9 \small$\pm{0.3} $ \\
					&CLIP-Adapter~\cite{pan_parameter-efficient_2022}  &     52.1\small$\pm{0.1}$\\
					 %\cline{2-5}
			         &CEFHRI-Adapter  &  73.3\small$\pm{0.3}$                \\
            \hline
			       \multirow{3}{*}	{HRI30}&  Adaptformer~\cite{chen_adaptformer_2022} &  60.7 \small$\pm{0.3} $ \\
					&CLIP-Adapter~\cite{pan_parameter-efficient_2022}  &     57.1\small$\pm{0.1}$\\
					 %\cline{2-5}
			         &CEFHRI-Adapter  &  80.6\small$\pm{0.1}$                \\
					\hline
		\end{tabular}}}
		}
		\label{tab:table5}
% \end{wraptable}
\end{table}

\begin{table}[h]
\caption{\footnotesize {Performance comparison with different \# of layers dropped. Here, $R=40$ , $N=16$, and $|\sS|=4$. All results represent the averaged percentage accuracy of HRI action recognition over several runs for $\alpha = 0.5$.}}
			
		\centering
  \scalebox{0.70}{
 		\resizebox{0.5\textwidth}{!}{
			{
				\begin{tabular}{l|c|c}
					\hline
	
					\textbf{Dataset}&\textbf{\bf Layers Dropped }              &    \bf Accuracy(\%)                                                                                   \\
				 \hline
					
				% \multirow{3}{*}	{COIN}&  Adaptformer~\cite{chen_adaptformer_2022} &  15.9 \small$\pm{0.3} $ \\
				% 	&CLIP-Adapter~\cite{pan_parameter-efficient_2022}  &     43.1\small$\pm{0.1}$\\
				% 	 %\cline{2-5}
			 %         &CEFHRI-Adapter  &  43.1\small$\pm{0.1}$                \\
				% 	 \hline
			       \multirow{4}{*}	{InHARD}&  1 &  72.3\small$\pm{0.3}$  \\
					&2  &     71.8\small$\pm{0.1}$\\
     &3  &     70.1\small$\pm{0.1}$\\
     &4  &     67.3\small$\pm{0.1}$\\
					 %\cline{2-5}
			         
            \hline
			       \multirow{4}{*}	{HRI30}&   1 &  79.8 \small$\pm{0.2} $ \\
					&2  &     79.1\small$\pm{0.1}$\\
     &3  &     77.5\small$\pm{0.1}$\\
     &4  &     72.1\small$\pm{0.1}$\\
					\hline
		\end{tabular}}}
		}
		\label{tab:table6}
% \end{wraptable}
\end{table}
\paragraph{Performance while Preserving 
Server Privacy}\label{server_Results}
In order to protect the privacy of the foundation model with $l$ layers,  we implement a model compression strategy by removing a specified number of layers, $n$ from the ViT-B/16~\cite{dosovitskiy_image_2021} model. The resulting compressed model is fine-tuned using knowledge distillation, with mean square error, under the supervision of the original model using Kinetics400~\cite{kay2017kinetics} dataset for 20 epochs. The fine-tuned compressed model is then distributed to the clients as a point of reference. The clients then use this reference model to train the $l-n$ adapters, which are later inserted into the server model's last $l-n$ layers. The results of our experiments indicate that knowledge distillation was essential in attaining the best performance. We also find that dropping the last $n$ layers produced superior results compared to dropping the first $n$ layers of the transformer model, which is consistent with~\cite{naseer2021intriguing}. Therefore, the reported results in Table~\ref{tab:table6} are generated using the strategy of dropping the last n layers for model compression. Here, we can observe that as the \# of removed layers  is more than 3, the performance drop is significant.

%\section{Discussion}
\section{Conclusion}
In this research, we present a new FL framework called CEFHRI, which aims to improve the performance of human-robot interaction recognition tasks in industrial settings through the utilization of pre-trained video models. To mitigate the communication overhead that is commonly encountered in FL systems, the CEFHRI framework proposes a parameter-efficient fine-tuning prototype. We conduct a comprehensive evaluation of the CEFHRI framework by comparing its performance against other baselines with regard to both recognition for human-robot interaction and communication cost. Our findings indicate that the CEFHRI framework not only effectively addresses the communication bottleneck issue but also outperforms other FL fine-tuning techniques, such as linear probing and bias-tuning. Additionally, our results show that the proposed CEFHRI adapter performs satisfactorily in scenarios where the downstream dataset has a major domain shift compared to the pre-trained dataset.  In conclusion, our work sheds new light on the potential for improving communication efficiency in FL for video understanding tasks and lays the groundwork for future advancements in this area.

\bibliographystyle{IEEEtran}
% \bibliography{intro,references,sim,rl_algo,musculoskeletal}
\bibliography{references}

\end{document}